\documentclass[runningheads]{llncs}

\usepackage{eccv}

\usepackage{eccvabbrv}

\usepackage{multirow}
\usepackage{mathtools}
\usepackage{array}
\usepackage{newfloat}
\usepackage{algorithmic}
\usepackage[linesnumbered,boxed,ruled,commentsnumbered]{algorithm2e}
\usepackage{setspace}

\usepackage{amsmath}
\usepackage{amssymb}
 
\usepackage{balance}

\usepackage{graphicx}
\usepackage{booktabs}

\usepackage[accsupp]{axessibility}  

\usepackage{hyperref}

\usepackage{orcidlink}

\begin{document}

\title{Semi-Supervised Teacher-Reference-Student Architecture for Action Quality Assessment} 

\titlerunning{Semi-Supervised Teacher-Reference-Student Architecture for AQA}

\author{Wulian Yun\orcidlink{0000-0003-0258-6044} \and Mengshi Qi\orcidlink{0000-0002-6955-6635}  \and Fei Peng\orcidlink{0009-0003-5035-9281} \and Huadong Ma\thanks{Corresponding author.}   \orcidlink{0000-0002-7199-5047}   
}

\authorrunning{W.~Yun et al.}


\institute{State Key Laboratory of Networking and Switching Technology, \\ Beijing University of Posts and Telecommunications, China
\email{\{yunwl,qms,pf0607,mhd\}@bupt.edu.cn}}

\maketitle

\begin{abstract}
Existing action quality assessment~(AQA) methods often require a large number of label annotations for fully supervised learning, which are laborious and expensive. In practice, the labeled data are difficult to obtain because the AQA annotation process requires domain-specific expertise. In this paper, we propose a novel semi-supervised method, which can be utilized for better assessment of the AQA task by exploiting a large amount of unlabeled data and a small portion of labeled data. 
Differing from the traditional teacher-student network, we propose a teacher-reference-student architecture to learn both unlabeled and labeled data, where the teacher network and the reference network are used to generate pseudo-labels for unlabeled data to supervise the student network. 
Specifically, 
the teacher predicts pseudo-labels by capturing high-level features of unlabeled data.
The reference network provides adequate supervision of the student network by referring to additional action information. Moreover, we introduce confidence memory to improve the reliability of pseudo-labels by storing the most accurate ever output of the teacher network and reference network. 
To validate our method, we conduct extensive experiments on three AQA benchmark datasets. Experimental results show that our method achieves significant improvements and outperforms existing semi-supervised AQA methods.

\end{abstract}

\section{Introduction}
Action quality assessment~(AQA) aims to assess~\emph{how well} the action is performed, which is an important task for video understanding and has been widely used for medical surgery skill assessment, sports activities scoring, etc. 
In recent years, many existing works~\cite{10.1007/978-3-319-10599-4_36,8014750,Parmar_2019_CVPR,8578732,9009513,an2024multi,10138608,10317826} have achieved significant progress on AQA task, which primarily relies on fully supervised learning and requires a large amount of annotated data for training. 
However, obtaining annotation labels for the AQA task is unique because the assessment process typically involves domain-specific knowledge, requiring annotators to have the relevant professional background and experience, resulting in a time-consuming and labor-intensive process of label acquisition.
Hence, it is important yet challenging to conduct better action assessments using very few labels.

\begin{figure}[t]
\begin{center}
\setlength{\fboxrule}{0pt}
\setlength{\fboxsep}{0cm}
\fbox{\rule{0pt}{0in} \rule{.0\linewidth}{0pt}
     \includegraphics[width=0.97\linewidth]{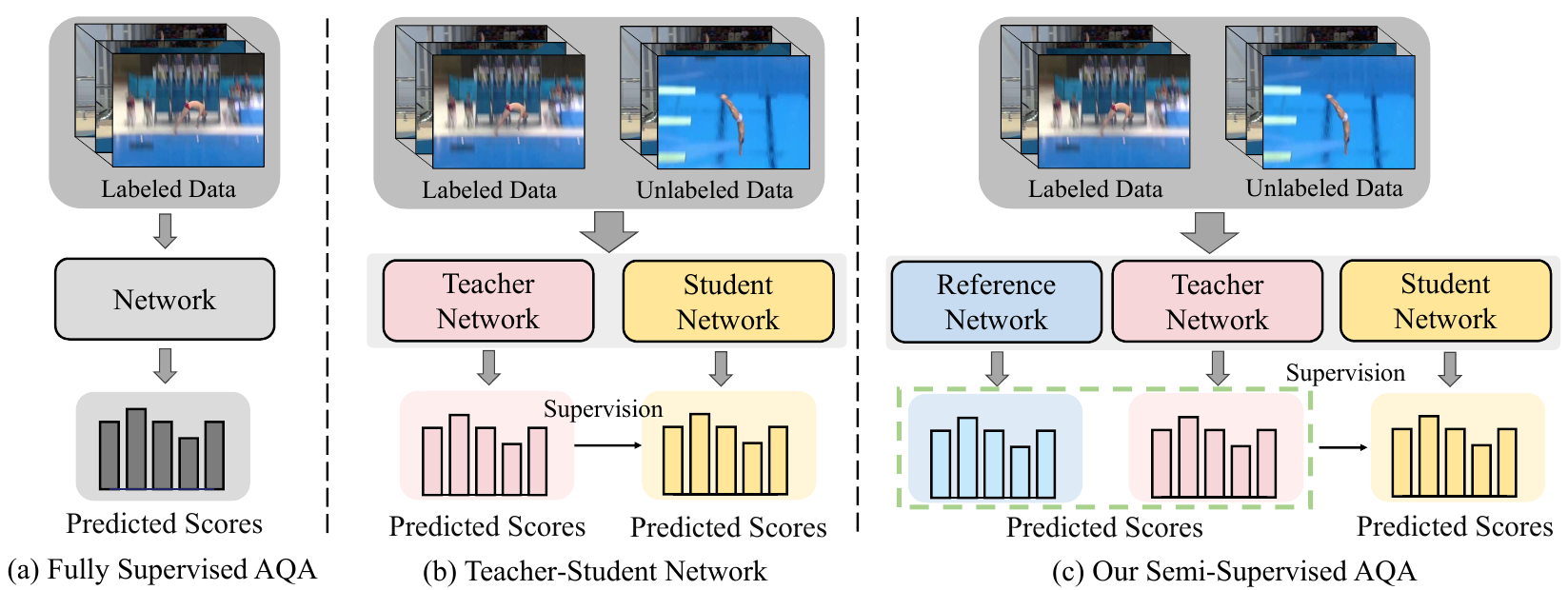}}
\end{center}
    \caption{Illustrations of (a) fully supervised AQA task, (b) traditional teacher-student network and (c) our semi-supervised AQA task. The fully supervised methods utilize labeled data as inputs for learning to achieve the assessment, while our semi-supervised method leverages labeled data and unlabeled data as inputs to perform the assessment.
    In addition, differing from the traditional teacher-student network, we have a reference network designed for the AQA task to provide supervision information for the student. }
\label{fig:1}
\end{figure}

Semi-supervised learning can effectively address the above issue by leveraging both labeled and unlabeled data for training,
thereby enhancing the performance of deep learning models.
Despite its significant advancements in various computer vision domains,~\emph{e.g.}, object detection~\cite{NEURIPS2019_d0f4dae8,Wang_2023_CVPR,9879982} and action recognition~\cite{10.5555/3454287.3454741,9879015}, there has been limited exploration of semi-supervised learning in the field of AQA.
Therefore, this motivates us to propose a semi-supervised approach for AQA, as shown in Figure~\ref{fig:1}. 
Specifically, we leverage the commonly employed teacher-student network~\cite{10.5555/3294771.3294885,9156610} in semi-supervised learning as the basis, in which the teacher generates pseudo-labels for unlabeled data to supervise the training of the student.

Ideally, the teacher network tends to predict more accurately results than the student network in semi-supervised learning, but there is no guarantee that the teacher's predictions invariably be better than the student's predictions. If the teacher generates low-fidelity pseudo-labels, it will affect the training of the student network and lead to poor assessment performance.
Especially for tasks like AQA that require learning subtle variations in the whole action, it remains a great challenge to generate high-quality pseudo-labels.

Inspired by the fact that teachers cannot always guarantee the infallibility of their insights when instructing students, and often need to refer to other knowledge and compare similarities and differences to ensure an improvement in teaching quality, we propose a novel semi-supervised teacher-reference-student architecture for AQA.
Specifically, our method mainly contains three networks, \ie, teacher network, student network and reference network.
The teacher network predicts quality scores by learning the high-level feature of each video, which first trains with labeled videos, and then generates pseudo-label for unlabeled videos to supervise the student network. 
The student network has the same structure as the teacher network.
The reference network predicts pseudo-labels by referring to other action information, thus providing additional supervision information for the student. 
In addition, considering that teachers have the memory capability, we construct a confidence memory to further improve the accuracy of pseudo-labels. The confidence memory stores the most accurate outputs ever predicted by the teacher network and reference network as output to supervise the student network, in which we introduce confidence to measure the reliability of pseudo-labels.

Our main contributions are summarized as follows:
\begin{itemize}
    \item We propose a novel semi-supervised  AQA method, which effectively leverages the knowledge from unlabeled and labeled data to improve the generalization of assessment.
    \item We introduce a reference network to provide additional supervision to the student network.  
    \item  We design a confidence memory to store the most accurate output predicted by the teacher network and reference network, which can improve the accuracy of pseudo-labels. 
    \item Extensive experimental results on three widely used datasets, \ie,  MTL-AQA, Rhythmic Gymnastics and JIGSAWS, demonstrate the effectiveness and superiority of our method.   
\end{itemize}

\section{Related Work}

\noindent
\textbf{Action quality assessment.}
Action quality assessment is an important task in video understanding ~\cite{yun2024weakly,qi2018stagnet,qi2020stc,qi2021semantics} areas. Existing mainstream AQA methods can be categorized into regression-based~\cite{10.1007/978-3-319-10599-4_36,Parmar_2019_CVPR,8014750,9009513,Tang_2020_CVPR,10.1007/978-3-031-19772-7_27} and ranking-based methods~\cite{8578732,Yu_2021_ICCV}. Regression-based methods mainly formulate the AQA as a regression task, by employing scores as supervised signals.
In early works, Pirisiavash~\emph{et al.}~\cite{10.1007/978-3-319-10599-4_36} utilize the Support Vector Machines~\cite{10.5555/2998981.2999003} to regress spatio-temporal pose features to scores. 
Pan~\emph{et al.}~\cite{9009513} learn the detailed joint motion by constructing graphs from both spatial and temporal aspects.
Tang~\emph{et al.}~\cite{Tang_2020_CVPR} adopt score distribution to depict the probability of the AQA score thereby addressing the problem of ambiguity of score labels.
Besides, Li~\emph{et al.}~\cite{10.1007/978-3-031-19772-7_27} propose a pairwise contrastive learning network to learn relative scores by capturing subtle and critical differences between videos. 
Subsequently, some works~\cite{8578732,Yu_2021_ICCV} formulate AQA as a ranking problem to further improve the accuracy of AQA.
However, the above methods rely on heavily samples with score labels, and obtaining these score labels is a time-consuming and labor-intensive process.
In contrast, our method focuses on achieving effective AQA performance with only a limited number of labeled samples. 

\noindent
\textbf{Semi-supervised learning.}
In recent years, semi-supervised learning has been widely used in various fields of computer vision, such as image classification~\cite{10.5555/3454287.3454741,NEURIPS2020_06964dce,9879015}, semantic segmentation~\cite{9879201,Wang_2022_CVPR}, object detection~\cite{NEURIPS2019_d0f4dae8,Wang_2023_CVPR,Liu_2022_CVPR,9879982}, etc.
Existing methods mainly divides into two branches, \ie consistency-based ~\cite{10.5555/3454287.3454741,10.5555/3294771.3294885,NEURIPS2020_06964dce,NEURIPS2021_995693c1,NEURIPS2019_d0f4dae8} and pseudo-label
based methods~\cite{9156610,Liu_2022_CVPR,9879982}. 
Consistency-based methods perturb the input data and leverage regularization loss to ensure that networks have consistent predictions for the same data under different augmentations.
For example, Meanteacher~\cite{10.5555/3294771.3294885} achieves semi-supervised learning by updating the parameters of the student to the teacher using the exponential moving average.
While pseudo-label based methods employ the teacher network to generate pseudo-label predictions for unlabeled data, and then leverage these pseudo-labels to optimize the student network.
A representative work is that NoisyStudent~\cite{9156610} utilizes the teacher to generate pseudo-labels for unlabeled data and introduces noise during the learning process to enhance the student's generalization ability.
However, so far semi-supervised learning is rarely explored in AQA. 
Zhang~\emph{et al.}~\cite{9682696} learn the temporal dependence of actions in unlabeled videos by self-supervised segment feature recovery for semi-supervised learning.
In contrast, we propose a novel semi-supervised teacher-reference-student architecture. This architecture utilizes the teacher network and reference network to generate pseudo-labels for unlabeled data, providing supervision for student training and facilitating comprehensive learning of unlabeled data.

\section{Preliminary}
\subsection{Teacher-Student Network}
Our semi-supervised architecture is based on a teacher-student network. The training process of the teacher-student network is mainly divided into two stages: \emph{burn-in stage} and \emph{teacher-student learning stage}. 

1) \emph{Burn-in stage}:~First, the teacher network is pre-trained on the labeled data in \emph{burn-in stage}, making the teacher network have enough regression ability to provide plausible pseudo-labels. The parameters of the teacher network are then used to initialize the student network. 

2) \emph{Teacher-student learning stage}:~During this stage, strong augmentation and weak augmentation are first performed on the input unlabeled data of the student network and the teacher network, respectively. This manner prevents the student network from overfitting based on excessive pseudo-labels. And then the teacher network supervises the training of the student network by making pseudo-label predictions for unlabeled data, where the parameters $\theta_t$ of the teacher network are updated from the student’s parameters $\theta_s$ via exponential moving average (EMA)~\cite{10.5555/3294771.3294885}, formulated as follows: 
\begin{equation}
\label{eq:1}
\theta_t=\alpha \theta_t+(1-\alpha) \theta_s ,
\end{equation}
where $\alpha \in(0,1)$ indicates the momentum.
This strategy can prevent the teacher network from overfitting on limited labeled data. Throughout the semi-supervised training, the student network is optimized with labeled and unlabeled data. The loss function optimizes the student network can be formulated as the following:
\begin{equation}
\label{eq:2}
L= L_{sup} + \beta L_{unsup} ,
\end{equation}
where $\beta$ is loss weight, $L_{sup}$ denotes the supervised loss and $L_{unsup}$ indicates the unsupervised loss. 
More details will be described in the following section.

\subsection{Task Definition}
In semi-supervised AQA, given a training set $D$ comprising a labeled subset $D^l=\left\{\left(v_i^l, s_i^l\right)\right\}_{i=1}^N$ and a unlabeled subset $D^u=\left\{v_i^u\right\}_{i=1}^M$. Specifically, $v_i^l$ and $v_i^u$ denote labeled video and unlabeled video, respectively. $s_i^l$ represents the score label of $v_i^l$, M and N refer to the number of unlabeled and labeled videos, 
where $M \gg N$. Semi-supervised AQA aims to simultaneously leverage the limited labeled data  $D^l$ as well as unlabeled data $D^u$ for training, and then
assess the quality score for each test video.

\section{Method}
\subsection{Pipeline}
\begin{figure*}[t]
\begin{center}
\setlength{\fboxrule}{0pt}
\setlength{\fboxsep}{0cm}
     \includegraphics[width=0.98\linewidth]{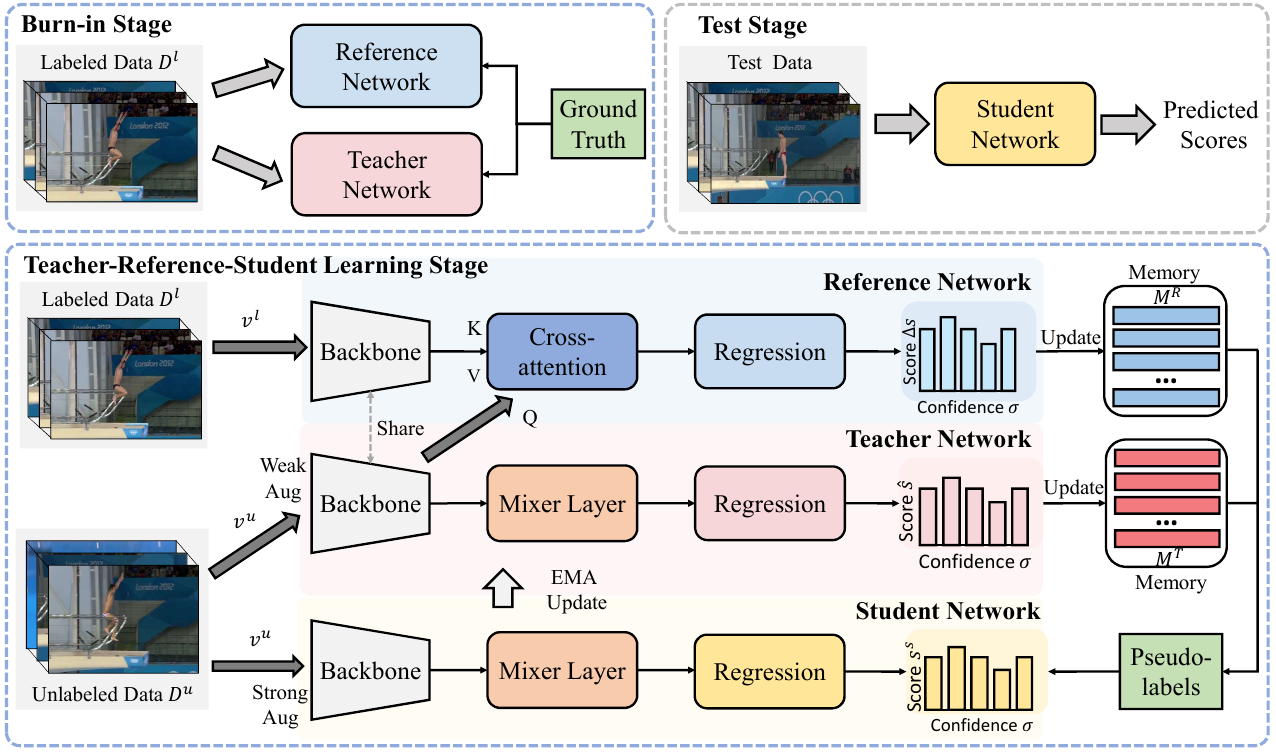}
\end{center}
\caption{
The overall framework of our method. Our method starts with a burn-in stage of training, followed by the teacher-reference-student learning stage. It mainly contains a teacher network, a student network and a reference network. 
The teacher network and reference network predict pseudo-labels for unlabeled data and supervise the training of the student network. The student network has the same architecture as the teacher.  
Moreover, confidence memory is used to ensure the reliability of pseudo-labels generated by the teacher network and reference network. 
}
\label{fig:2}
\end{figure*}

The overall framework of our method is illustrated in Figure~\ref{fig:2}. Specifically, our method consists of three main parts, \ie, teacher network, student network and reference network. 
The teacher network learns the quality score of each video. The student network and the teacher network have the same architecture, differing mainly in how the weights are updated. 
The reference network learns the relative scores between different videos to provide additional supervision information for the student.

The training process of our method is mainly as follows: 1) 
the teacher network and reference network will use labeled videos $v_p^l$ and $v_q^l$ from labeled data $D^l$ for a short period of training during the \emph{burn-in stage}, enabling both the teacher network and the reference network with the ability to assess quality scores. 
To be specific, the teacher network takes $v_p^l$ as input and directly estimates the score $\hat{s}_p^l$ of $v_p^l$. The reference network takes a pair of videos $<v_p^l, v_q^l>$ as input and estimates a relative score $\Delta \mathrm{s}^l$ between $v_p^l$ and $v_q^l$.
Then, the parameters $\theta_t$ of the teacher network are used to initialize the student network $\theta_s$.

2) Followed by the teacher-student learning stage, which in our process we call the \emph{teacher-reference-student learning stage}. The teacher network and the reference network are used to generate pseudo-labels for unlabeled data $D^u$, thus supervising the training of the student network. 
In this process, our method takes the unlabeled data $v_i^u$ as well as the video $v_i^l$ with known quality score $s_i^l$ from labeled data $D^l$, where $v_i^u$ is the input to the teacher network and $v_i^u$,   $v_i^l$ sever as the input to the reference network. 
We perform strong augmentation (\ie, Gaussian, and RandomHorizontalFlip) on the unlabeled data to the student network and weak augmentation (\ie, RandomHorizontalFlip) to the teacher network.
Meanwhile, we introduce confidence memory to guarantee the reliability of pseudo-labels generated by the teacher network and the reference network, to ensure that the pseudo-label output by networks is high-fidelity.
The scores of memory in the teacher network and the reference network are averaged as the final pseudo-label $\bar{\mathrm{s}}$ to supervise the student. 
The teacher’s parameters $\theta_t$ are updated by EMA of the student’s parameters  $\theta_s$ as presented in Eq.\eqref{eq:1}.
Next, we will present details about each component.

\subsection{Teacher Network and Reference Network }
We take the teacher-reference-student learning stage as an example to develop a detailed description of the teacher network and the reference network.

\noindent
\textbf{Teacher network} aims to directly map the unlabeled video $v^u$ to obtain the quality score $\hat{\mathrm{s}}^u$ based on the performance of human action and then supervise student's training. It can be defined as:
\begin{equation}
\label{eq:3}
\hat{\mathrm{s}}^u = \mathbf{A} (v^u),
\end{equation}
where $\mathbf{A} (\cdot)$ denotes the teacher network.
Specifically, for the video $v^u$, we first follow~\cite{Yu_2021_ICCV,10.1007/978-3-031-19772-7_25} to split the video into multiple overlapping snippets, where each snippet contains $C$ consecutive frames. Then, I3D~\cite{Carreira_2017_CVPR} is employed as the backbone to extract the feature of each snippet, thus obtaining video feature representation $F^u$.
After that, we utilize the Mixer Layer~\cite{NEURIPS2021_cba0a4ee} to extract higher-level stable features by encoding the temporal information of the video sequence. 
The Mixer Layer is comprised of~\emph{token-mixing} MLP and~\emph{channel-mixing} MLP, and each MLP contains two fully connected layers and a GELU~\cite{hendrycks2016gelu}. 
Concretely, we leverage the feature $F^u$ as input for Mixer Layer. $F^u$ is first fed into~\emph{token-mixing} MLP for interaction between snippets at different temporal positions, and then the feature $F^t$ output by~\emph{token-mixing} MLP are transposed and sent to~\emph{channel-mixing} MLP for independent interactions in each snippet dimension.
The process can be formulated as: 
\begin{equation}
\label{eq:4}
\begin{array}{ll}
{{F}^t} = F^u + f_2^t  \boldsymbol{g} (f_1^t (\text {LN}(F^u) )), \\
{{F}^c} = F^t + f_2^c    \boldsymbol{g} (f_1^c (\text {LN}({F}^t) )),
\end{array}
\end{equation}
where $\mathrm{LN}(\cdot)$ represents the LayerNorm~\cite{DBLP:journals/corr/BaKH16} operation, $\boldsymbol{g}$ denotes GELU, $f_1^t$, $f_2^t$ and $f_1^c$, $f_2^c$ denote the learnable weight matrices of fully-connected layers in~\emph{token-mixing} MLPs and \emph{channel-mixing} MLPs, respectively. 
Finally, the feature ${F}^c$ generated by Mixer Layers is fed into the regression layer to obtain the quality score $\hat{\mathrm{s}}^u$  of the teacher network. 

\noindent
\textbf{Reference network} aims to refer to other action information to improve the regression learning capability of the network and the accuracy of pseudo-labels. 
The reference network employs the attention mechanism to capture differences and similarities between videos, generating pseudo-labels and providing additional supervision information for the student network.
To be specific, given an video $v^l$ with known quality score  $s^l$ and an unlabeled video $v^u$, the reference network aims to estimate the relative score $\Delta \mathrm{s}^u$ of $<v^u, v^l>$:
\begin{equation}
\label{eq:5}
 \Delta \mathrm{s}^u =\Upsilon\left(v^u, v^l\right) ,
\end{equation}
where $\Upsilon (\cdot)$ denotes the reference network, video $v^l$ comes from the labeled set $D^l$, $v^u$ comes from unlabeled set $D^u$.

In detail, the reference network first performs feature extraction on video $v^l$ by backbone I3D to obtain feature $F^l$.
Subsequently, we utilize cross-attention learning to learn the relative relationship by distinguishing subtle differences in the temporal level of video pair $<v^u, v^l>$.
The feature $F^u$ of $v^u$ is treated as a query, and the feature $F^l$ of $v^l$ is served as a key and value, respectively. The learning process can be formulated as follows:
\begin{equation}
\label{eq:6}
\begin{aligned}
\operatorname{Attention}(\cdot) & =\operatorname{Softmax}\left(Q K^T / \sqrt{d_k} \right) V, \\
F^{\prime} & =\operatorname{Attention} (\operatorname{LN} (F^u, F^l ) )+F^u, \\
F^{\prime \prime} & =\operatorname{MLP}\left(\operatorname{LN} (F^{\prime}\right) )+F^{\prime},
\end{aligned}
\end{equation}
where $Q, K, V$ are the query, key and value matrices, $\mathrm{LN}(\cdot)$ denotes the LayerNorm operation,  and $d$ is the dimension of query and key features.
Finally, we predict the score $\Delta \mathrm{s}^u$ through the regression layer based on feature $F^{\prime \prime}$ of interaction between the video pair $<v^u, v^l>$.

\subsection{Confidence Memory}
\noindent
To further enhance the accuracy of pseudo-labels, we take into account that teachers have memory capability and thus want to utilize memory to store the best output of the teacher network and reference network. However, how to measure whether the network prediction results are reliable is a key issue. Consider that in object detection\cite{10.1007/978-3-031-25085-9_2,8953889} and pose estimation\cite{9710108,8953904} tasks, the reliability of outputs usually relies on the confidence measure. Therefore, we propose a confidence memory that stores the best-reliable score output of the teacher network and the reference network by confidence evaluation during the teacher-reference-student learning stage.

Specifically, we set an empty confidence memory $M^t$, which is used to store the quality score and confidence of each unlabeled data predicted by the teacher network. 
In each training iteration, the teacher network retrieves videos’ confidence $\boldsymbol{\sigma}$ and quality score from memory using a read operation,
if the confidence $\boldsymbol{\sigma}$ of the current video prediction is higher than the confidence stored in memory $M^t$, the score of the current prediction is archived into memory $M^t$ by a write operation. Simultaneously, the confidence $\boldsymbol{\sigma}$ in memory $M^t$ is also updated.
Regarding the confidence estimation, we leverage a single variate Gaussian following~\cite{8953904,8953889} to generate the probability distribution $P_\Theta(\mathbf{x})$ of the predicted score:
\begin{equation}
\label{eq:7}
P_{\Theta}(\mathbf{x})=\frac{1}{\sqrt{2 \pi}  {\boldsymbol{\sigma}}} \exp \left(-\frac{(\mathbf{x}- {\boldsymbol{\mu}})^2}{2 {\boldsymbol{\sigma}}^2}\right)  ,
\end{equation}
where $\Theta$ represents learnable parameters,
$\boldsymbol{\mu}$ denotes the predicted score of the teacher network. 
$\boldsymbol{\sigma}$ denotes the standard deviation, which is used as a measure of prediction uncertainty. When $\boldsymbol{\sigma} \rightarrow 0$, our network is confident in the predicted result.
In this way, we construct another confidence memory bank $M^r$ to store high-fidelity pseudo-labels predicted by the reference network. It is worth noting that the predicted scores stored in memory $M^t$ are $\hat{\mathrm{s}}^u$, and the predicted scores in memory $M^r$ are $\Tilde{\mathrm{s}}^u =\mathrm{s}^l+\Delta \mathrm{s}^u$.

\subsection{Optimization and Inference}
The final optimization objective of our method contains both supervised loss $\mathcal{L}_{sup}$ and unsupervised loss $\mathcal{L}_{unsup}$.
The supervised loss $\mathcal{L}_{sup}$ is mainly composed of $\mathcal{L}_{reg}^s $ and $ \mathcal{L}_{reg}^r$. The main goal of $\mathcal{L}_{reg}^s $ is to minimize the error between the score $\mathrm{s}^s$ predicted by the student network and the ground truth $\mathrm{s}$, and $ \mathcal{L}_{reg}^r$ aims to minimize the error between the predicted relative score $\Delta \mathrm{s}$ of reference network and the corresponding ground truth $\mathrm{s}$, $\mathrm{s}^l$. The loss function of $\mathcal{L}_{sup}$ can be defined as:
\begin{equation}
\label{eq:8}
\begin{aligned}
\mathcal{L}_{sup} & = \mathcal{L}_{reg}^s + \mathcal{L}_{reg}^r, \\
\mathcal{L}_{reg}^s & = -\log P_\Theta(\mathbf{x})\Big|_{\mathbf{x} =\mathrm{s}}\propto\log\hat{\boldsymbol{\sigma}}+\frac{(\mathrm{s} -\mathrm{s}^s)^2}{2\hat{\boldsymbol{\sigma}}^2}  ,\\
\mathcal{L}_{reg}^r & = -\log P_\Theta(\mathbf{x})\Big|_{\mathbf{x} =\Delta \mathrm{s}}\propto\log\hat{\boldsymbol{\sigma}}+\frac{(\Delta \mathrm{s} -\left| \mathrm{s}-\mathrm{s}^l\right|)^2}{2\hat{\boldsymbol{\sigma}}^2}  .\\
\end{aligned}
\end{equation}

The unsupervised loss $\mathcal{L}_{unsup}$ can be formulated as:
\begin{equation}
\label{eq:9}
\mathcal{L}_{unsup} = -\log P_\Theta(\mathbf{x})\Big|_{\mathbf{x} =\bar{\mathrm{s}}}\propto\log\hat{\boldsymbol{\sigma}}+\frac{(\bar{\mathrm{s}}-\mathrm{s}^s)^2}{2\hat{\boldsymbol{\sigma}}^2} ,
\end{equation}
where $\mathrm{s}^s$ denote the quality score predicted by the student network,
$\bar{\mathrm{s}}$ is the final pseudo-label provided by the teacher network and reference network.
Finally, the optimization object for our method can be summarized as:
\begin{equation}
\label{eq:10}
L_{total}= L_{sup} + \beta L_{unsup},
\end{equation}
where $\beta$ denotes loss weight. During the inference stage, we leverage only the student network to predict quality scores for test videos. Overall, the complete training process of our method is described in Algorithm~\ref{alg:1}. 

\begin{algorithm}[H]
\setstretch{0.7}
\small
 \caption{The training process of our method.}
 \label{alg:1}
 \LinesNumbered  
  \KwIn{Labeled data ${(v^l,s^l)\in D^l}$, unlabeled data $v^u\in D^u$, burn-in step $b$, maximum iteration $E$.}
 \KwOut{Teacher model $\theta_t$, Student model $\theta_s$.}
 \For{$e < E$}{\If{ $e < b$}{Predict scores $\hat{\mathrm{s}}^l$ and $\Delta \mathrm{s}^l$ of  $v^l$ by Teacher $\theta_t$ and Reference $\theta_f$  \; 
 Calculate $L_{sup}$ and Update $\theta_t$, $\theta_f$ by $\mathcal{L}_{sup}$ in Eq.\eqref{eq:8} \;}
 {\If{ $e = b$}{Initialize $\theta_s$ with $\theta_t$ }}
 {\If{ $e > b$}{Predict pseudo-label $\bar{s}$ of $v^u$ by Teacher $\theta_t$ and Reference $\theta_f$ \;
 Predict score $s^s$ of $v^u$ by Student $\theta_s$  \;
  Calculate $L_{unsup}$ with $\left(\bar{s}, s^s\right)$ \;
 Calculate $L_{sup}$ with $D^l$ \;
  Update $\theta_s$ and $\theta_f$ by Eq.\eqref{eq:10} and Eq.\eqref{eq:8} \; 
  Update $\theta_t$ by $\theta_s$ with EMA in Eq.\eqref{eq:1}}
}}
Return $\theta_t$ and $\theta_s$
\end{algorithm}

\section{Experiments}
\subsection{Datasets}
We conduct our experiments on the three commonly-used benchmark datasets, including MTL-AQA~\cite{Parmar_2019_CVPR}, Rhythmic Gymnastics~\cite{zeng2020hybrid} and JIGSAWS~\cite{gao2014jhu}.

\noindent
\textbf{MTL-AQA} contains 1,412 diving videos, where videos are collected from 16 different international competitions. 
Following the previous work~\cite{Tang_2020_CVPR,Yu_2021_ICCV,10.1007/978-3-031-19772-7_27}, we use 1,059 training videos to train our model and 353 test videos for evaluation.

\noindent
\textbf{Rhythmic Gymnastics} comprises 1,000 videos from international competitions, consisting of four gymnastics routines, \emph{i.e.}, ball, ribbon, hoop, and clubs. Each routine has 250 videos, each of which is approximately 1.6 minutes in length and has a frame rate of 25 frames.
Following~\cite{zeng2020hybrid,9682696}, 
we train separate models for each kind, with 200 videos used for training and 50 videos used for evaluation.

\noindent
\textbf{JIGSAWS} is a surgical activity dataset, which consists of 103 videos from three different types of surgical activities: Suturing, Needle Passing and Knot Tying. To assess different aspects of surgical actions, each video sample is annotated with multiple scores and the final score is the sum of all annotations.

\subsection{Evaluation Metric}
Following~\cite{Tang_2020_CVPR,Yu_2021_ICCV,10.1007/978-3-031-19772-7_27}, we exploit the widely-adopted Spearman’s rank correlation~(Sp.Corr) $\rho$ as evaluation metric to evaluate the performance of our method. Spearman’s rank correlation mainly calculates the difference between ground-truth scores and predicted scores, which ranges from -1 to 1 and the higher value of $\rho$ indicates better performance.
$\rho$ can be defined as:
\begin{equation}
\rho=\frac{\sum_i\left(x_i-\bar{x}\right)\left(y_i-\bar{y}\right)}{\sqrt{\sum_i\left(x_i-\bar{x}\right)^2 \sum_i\left(y_i-\bar{y}\right)^2}}
\end{equation}
where $x$ and $y$ denote the ranking of two series, respectively.

\subsection{Implementation Details}
We implement our method with PyTorch \cite{paszke2019pytorch} and all models are trained by Adam optimizer~\cite{2014Adam}. 
We utilize the I3D~\cite{Carreira_2017_CVPR} backbone pre-trained on  Kinetics~\cite{DBLP:journals/corr/KayCSZHVVGBNSZ17} as the feature extractor.
The total training epoch is set as 150. 
For MTL-AQA~\cite{Parmar_2019_CVPR} dataset, 
we sample 103 frames for each video following~\cite{Yu_2021_ICCV}
and segment them into 10 overlapping snippets each containing 16 continuous frames with a stride of 10 frames. The learning rate is 2e-4. 
For Rhythmic Gymnastics~\cite{zeng2020hybrid} dataset, following previous work~\cite{9682696}, we uniformly sample 2 frames per second from each original video, resulting in a sampled video of about 180 frames. Then, we divide the sampled video into 18 snippets. The learning rate is set to 4e-4. 
For JIGSAWS~\cite{gao2014jhu} dataset, 
160 frames are extracted from each video, which are divided into 10 overlapping snippets, each containing 16 consecutive frames.
The learning rate is 1e-4. Four-fold cross-validation is performed during the experiment by following~\cite{Tang_2020_CVPR,9682696}.
All the video frames are resized to 224 $\times$ 224.
The weight $\beta$ of unsupervised loss is updated with training
epoch $t$ following an exponential warming-up function~\cite{liu21DMT-Net}:
$\lambda(t)=0.2 \times e^{-5(1-t / 200)^2}$.

\subsection{Comparison with State-of-the-Art Methods}
We compare our method with existing state-of-the-art AQA methods on three public benchmarks, the results are shown in Table \ref{tab:1}, Table \ref{tab:2} and Table \ref{tab:3}.

\noindent
\textbf{Results on MTL-AQA dataset.}
We first compare our method with SOTA fully supervised methods~(\ie, SVR~\cite{10.1007/978-3-319-10599-4_36}, C3D-AVG-STL~\cite{Parmar_2019_CVPR}, C3D-AVG-MTL~\cite{Parmar_2019_CVPR}, USDL~\cite{Tang_2020_CVPR}, CoRe~\cite{Yu_2021_ICCV} and TPT~\cite{10.1007/978-3-031-19772-7_25}) and semi-supervised methods(\ie, COREG~\cite{10.5555/1642293.1642439}, Pseudo-labels~\cite{hou2017semi}
 VAT~\cite{Adversarial}, $S^4$L~\cite{9010283} and $S^4$AQA~\cite{9682696}) on MTL-AQA dataset.
Our method utilizes the student network to directly predict scores without additional data.
In contrast, CoRe and TPT employ a complex multi-exemplar voting strategy that requires multiple additional training data to be introduced for testing. To ensure fairness, we adapt their testing strategy to randomly introduce a video from the training set classified as unlabeled for testing, and randomly select video for training.
As shown in Table~\ref{tab:1}, our method can obtain obvious gains over existing semi-supervised methods while also surpassing fully supervised methods.
On the MTL-AQA dataset with $40\%$ labels, our method outperforms the results of semi-supervised method $S^4$AQA~\cite{9682696} by 20.8\%. Also, compared with $S^4$AQA on only $10\%$ label, the performance gains by our method are significant, \ie, + 22.0\%. 
In summary, the results demonstrate that our method can sufficiently utilize both unlabeled and labeled data to improve the performance of AQA.

\noindent
\textbf{Results on Rhythmic Gymnastic dataset.}
Our method achieves state-of-the-art results in four categories of the Rhythmic Gymnastic dataset with $40\%$ of labeled data, shown in Table~\ref{tab:2}. We can find that our method outperforms semi-supervised method $S^4$AQA~\cite{9682696} by 0.187 and fully supervised method ACTION-NET~\cite{zeng2020hybrid} by 0.211, demonstrating the effectiveness of our method.

\noindent
\textbf{Results on JIGSAWS  dataset.}  
Table~\ref{tab:3} shows the evaluation results on JIGSAWS dataset with $50\%$ of labeled data. From the table, our model achieves competitive performance compared to other methods. 
This shows that we can achieve better results not only in the sports domain but also in the surgical domain, proving the generalizability of our method.

\begin{table}[t]
\caption{Comparison with existing AQA methods on MTL-AQA dataset. $^*$ denotes the fully supervised method. The best results are highlighted in bold.}
\label{tab:1}
\centering
\renewcommand{\arraystretch}{1.0}
\scalebox{0.79}{
 \setlength{\tabcolsep}{7.5mm}{
\begin{tabular}{cccc}
\specialrule{0.1em}{0.5pt}{0.5pt}
\hline
\multirow{2}{*}{Methods} & \multicolumn{2}{c}{Semi-supervision} \\ \cline{2-3} 
                         & 10\%         & 40\%        \\ \hline
  SVR$^*$~\cite{10.1007/978-3-319-10599-4_36}          &  0.427            &      0.565       \\ 
 
  C3D-AVG-STL$^*$~\cite{Parmar_2019_CVPR}            &  0.561            &      0.632    \\ 
  C3D-AVG-MTL$^*$~\cite{Parmar_2019_CVPR}            &  0.584            &      0.656       \\
  USDL$^*$~\cite{Tang_2020_CVPR}                   &  0.530            &    0.646       \\
  CoRe$^*$~\cite{Yu_2021_ICCV}      & 0.773   &  0.894  \\
  TPT$^*$~\cite{10.1007/978-3-031-19772-7_25}                   &  0.723            &    0.781       \\
  \hline
  COREG~\cite{10.5555/1642293.1642439}                 &  0.487            &      0.526      \\
  Pseudo-labels~\cite{hou2017semi}        &  0.622            &      0.716       \\
  VAT~\cite{Adversarial}                    &  0.635            &      0.724      \\
  $S^4$L~\cite{9010283}                 &  0.621            &      0.721        \\
  $S^4$AQA~\cite{9682696}            &  0.676            &      0.746       \\\hline
  \textbf{Ours}          &   \textbf{0.825}          &    \textbf{0.901} \\
  \hline\specialrule{0.1em}{1pt}{1pt}
\end{tabular}
}}
\end{table}

\begin{table}[ht]
\caption{Comparison with existing AQA methods on Rhythmic Gymnastic dataset. $^*$ denotes the fully supervised method. The best results are highlighted in bold.}
\label{tab:2}
\centering
\renewcommand{\arraystretch}{0.95}
\scalebox{0.80}{
\setlength{\tabcolsep}{3.15mm}{
\begin{tabular}{cccccc}
\specialrule{0.1em}{0.5pt}{0.5pt}
\hline
  \multirow{2}{*}{Methods} & \multicolumn{5}{c}{Rhythmic Gymnastic( 40\%  )} \\ \cline{2-6} 
                        & Ball  & Clubs & Hoop  & Ribbon    &AVG   \\ \hline 
  SVR$^*$~\cite{10.1007/978-3-319-10599-4_36}              &  0.175     &  0.243  & 0.261  &  0.309  &  0.248    \\ 
  ACTION-NET$^*$~\cite{zeng2020hybrid}      &  0.196       & 0.403     & 0.319   & 0.305  & 0.308       \\ 
 \cline{1-6} 
  COREG~\cite{10.5555/1642293.1642439}                 &  0.230            &  0.338   & 0.331  & 0.268 & 0.292         \\
  Pseudo-labels~\cite{hou2017semi}        &  0.183            &  0.330  & 0.346   & 0.305 &  0.292        \\
  VAT~\cite{Adversarial}     &  0.208    &  0.355   & 0.345  & 0.292 & 0.301       \\
  $S^4$L~\cite{9010283}       &  0.209   &  0.325   & 0.324  & 0.290  &  0.288    \\
 $S^4$AQA~\cite{9682696}      &  0.248   &  0.388    & 0.307  & 0.357 & 0.342   \\ \hline
  \textbf{Ours}          &   \textbf{0.558}         &  \textbf{0.518}    &  \textbf{0.567} & \textbf{0.475}   & \textbf{0.529} \\ 
  \hline\specialrule{0.1em}{1pt}{1pt}
\end{tabular}
}}
\end{table}

 \begin{table}[ht]
 \caption{Comparison with existing AQA methods on JIGSAWS dataset. $^*$ denotes the fully supervised method. The best results are highlighted in bold.}
\label{tab:3}
\centering
\renewcommand{\arraystretch}{0.95}
\scalebox{0.80}{
\setlength{\tabcolsep}{2.9mm}{
\begin{tabular}{ccccc}
\specialrule{0.1em}{0.5pt}{0.5pt}
\hline
  \multirow{2}{*}{Methods} & \multicolumn{4}{c}{JIGSAWS( 50\%  )} \\ \cline{2-5} 
                        & Suturing  & Needle Passing & Knot Tying     &AVG   \\ \hline 
  USDL$^*$~\cite{Tang_2020_CVPR}                    &  0.439     &  0.351  & 0.680  &  0.505   \\ 

 CoRe$^*$~\cite{Yu_2021_ICCV}      & 0.622   & 0.776  & 0.753 &  0.717 \\ \hline
  
  Pseudo-labels~\cite{hou2017semi}        &  0.445            &  0.501  & 0.714   & 0.566       \\
  VAT~\cite{Adversarial}     &  0.524    &  0.526   & 0.749  & 0.612    \\
  $S^4$L~\cite{9010283}       &  0.455   &  0.529   & 0.730  & 0.585    \\
 $S^4$AQA~\cite{9682696}      &  0.533   &  0.552    &  \textbf{0.813}  & 0.655   \\ \hline
  \textbf{Ours}          &   \textbf{0.696}         &  \textbf{0.810}    &   0.753  & \textbf{ 0.753 }    \\ 
  \hline\specialrule{0.1em}{1pt}{1pt}
\end{tabular}
}}
\end{table}

\begin{table}[t]
\caption{Ablation study on each component of our method on MTL-AQA dataset. `Base' indicates baseline, `RN' denotes the reference network, and `TM' and `RM' represent confidence memory of the teacher network and the reference network, respectively. }
\label{tab:5}
\centering
\renewcommand{\arraystretch}{0.95}
\scalebox{0.79}{
\setlength{\tabcolsep}{3.3mm}{
\begin{tabular}{lccccc}
\specialrule{0.1em}{0.5pt}{0.5pt}
\hline
\multicolumn{1}{c}{\multirow{2}{*}{Method}} & \multicolumn{1}{c}{\multirow{2}{*}{RN}} & \multicolumn{1}{c}{\multirow{2}{*}{TM}} & \multicolumn{1}{c}{\multirow{2}{*}{RM}} & \multicolumn{2}{c}{MTL-AQA}  \\ \cline{5-6} 
\multicolumn{1}{c}{}                        & \multicolumn{1}{c}{}                    & \multicolumn{1}{c}{}                     & \multicolumn{1}{c}{}                     & \multicolumn{1}{c}{10\%} &  \multicolumn{1}{c}{40\%} \\ \hline
Base                                         &                                          &                                           &                                           & \multicolumn{1}{c}{ 0.751} &  0.865      \\ 
Base + TM                                      &                         &       \; $\checkmark$                                      &                                           & \multicolumn{1}{c}{0.794}        &   0.870        \\ 
Base + RN                                     &  \; $\checkmark$                         &                                           &                                          & \multicolumn{1}{c}{0.811 }        &  0.891       \\ 
Base + RN + TM                              &       \; $\checkmark$                                     &     \; $\checkmark$                       &                                       & \multicolumn{1}{c}{ 0.813}           &  0.893        \\ \hline
\textbf{Ours}                                         &    \; $\checkmark$                                        &     \; $\checkmark$                       &    \; $\checkmark$                        & \multicolumn{1}{c}{ \textbf{0.825}}          &    \textbf{0.901}     \\ \hline

  \specialrule{0.1em}{1pt}{1pt}
\end{tabular}
}}
\end{table}

\begin{figure}[ht]
\centering
\setlength{\fboxrule}{0pt}
\setlength{\fboxsep}{0cm}
     \includegraphics[width=0.75\linewidth]{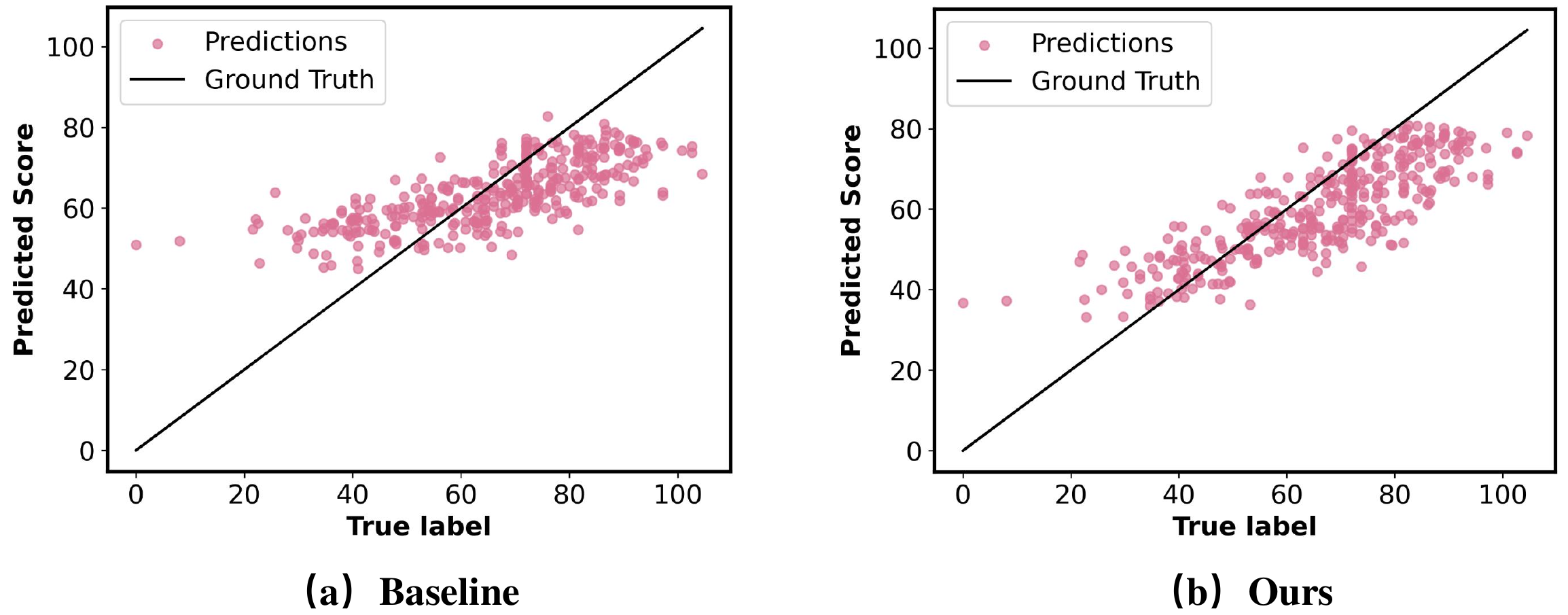}
\caption{Scatter plot comparison results of score prediction between baseline and our method on MTL-AQA dataset. The black line indicates the ground truth, while the pink points represent the predicted scores.}
\label{fig:3}
\end{figure}

\begin{table}[ht]
\caption{The experiment results of different proportions of labeled data.}
\label{tab:6}
\centering
\renewcommand{\arraystretch}{0.95}
\scalebox{0.79}{
\setlength{\tabcolsep}{5.8mm}{
\begin{tabular}{ccccc}
\specialrule{0.1em}{0.5pt}{0.5pt}
\hline
\multirow{2}{*}{Methods} & \multicolumn{4}{c}{MTL-AQA} \\ \cline{2-5}     
    & 10\%     & 20\%  & 30\%   & 40\%      \\ \hline 
Ours             &   0.825 & 0.851 &  0.862        &    0.901    \\
  \hline
  \specialrule{0.1em}{1pt}{1pt}
\end{tabular}
}}
\end{table}

\begin{figure*}[ht]
\begin{center}
\setlength{\fboxrule}{0pt}
\setlength{\fboxsep}{0cm}
     \includegraphics[width=0.9\linewidth]{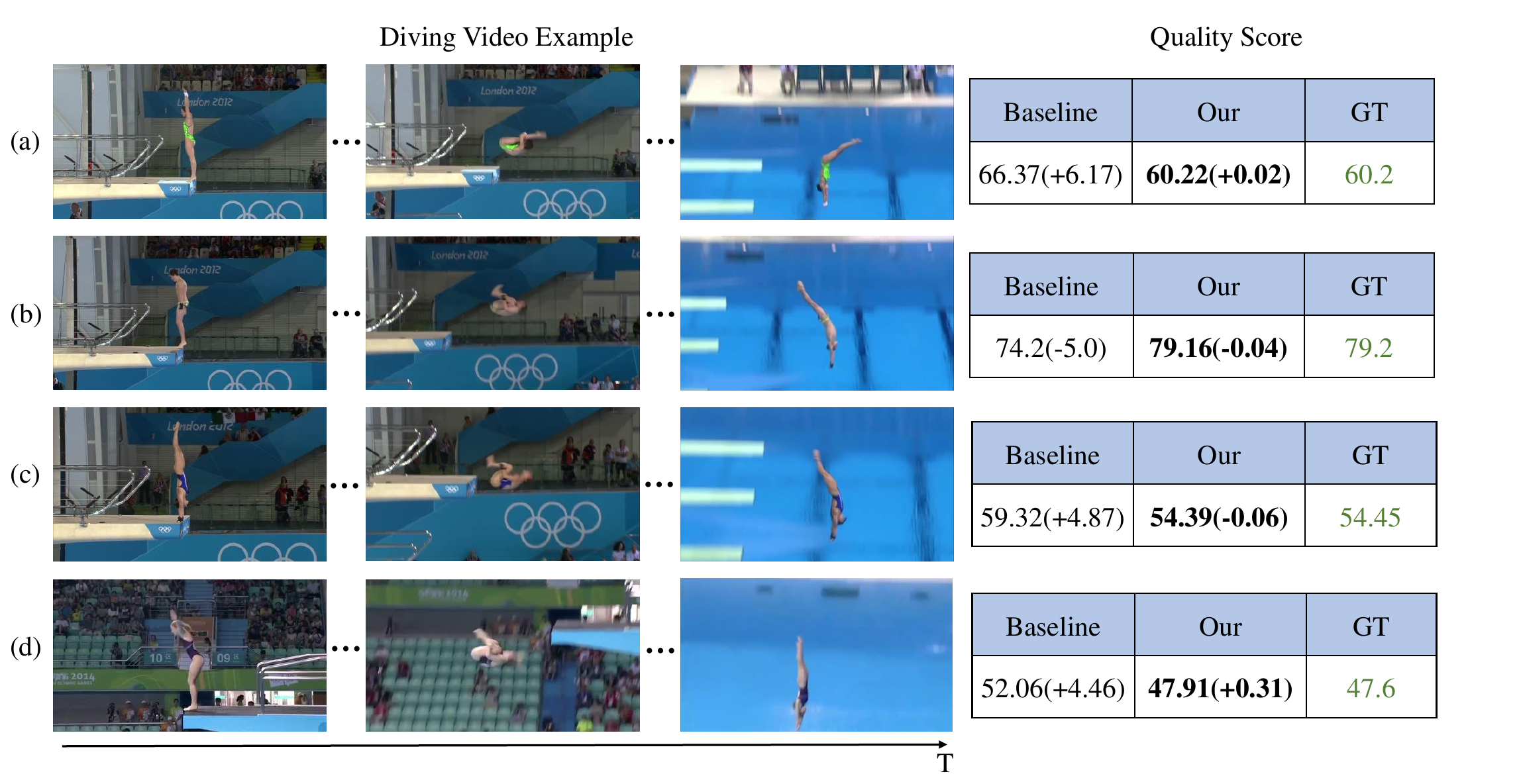}
\end{center}
\caption{
Case study with qualitative results on four video samples from the MTL-AQA dataset, which presents the predicted scores comparisons of baseline, our method and ground truth~(GT). `Baseline' means the traditional teacher-student network. 
}
\label{fig:4}
\end{figure*}

\subsection{Ablation Study}
\noindent
\textbf{The impact of different components.}
We evaluate the effect of different components of our method, the results are shown in Table~\ref{tab:5}.
We set up the traditional teacher-student network in Sec 3.1 as the baseline, and then gradually add confidence memory, and reference network to the baseline. Where 1) \emph{Base} denotes the traditional teacher-student network; 2) \emph{Base+TM} denotes add confidence memory into the teacher network of 1); 3) \emph{Base+RN} denotes add a reference network into 1); 4) \emph{Base+RN+TM} denotes add confidence memory in teacher network of 3);  5) \emph{Ours} denotes add confidence memory into the reference network of 4), \ie, our full model.
From the results, it can be observed that with the increase of different components, the assessment results gradually increase with the use of different proportions of labeled data, which is enough to prove the importance of different components.

\noindent
\textbf{The impact of memory updating strategy.}
To validate the effectiveness of the updating memory by confidence measure in our method, we explore another memory updating strategy, \ie, direct updating based on the output of the current epoch, and the experimental results are shown in \emph{Base+RN} of Table~\ref{tab:5}. We can find the direct update strategy underperforms the model with the confidence-based updating (\ie, \emph{Base+RN+TM}). The results can prove that updating the memory according to the confidence measure is effective. 

\noindent
\textbf{Sensitivity to different proportions of labeled data.}
To verify the sensitivity of our method when using different proportions of labeled data, we conducted experiments with $10\%, 20\%, 30\%$ and $40\%$ of labeled data, and the results are shown in Table~\ref{tab:6}.
As seen from the table, the prediction performance of our method gradually increases with the increase of labeled data. Meanwhile, we have found that our proposed method with only $10\%$ of labeled data outperforms the $S^4$AQA~\cite{9682696} with $40\%$ labeled data in Table~\ref{tab:1}, and these findings are sufficient to show that our proposed method can effectively learn the rich representations from the unlabeled video.

\subsection{Qualitative Analysis}

To further intuitively prove the effectiveness of our proposed method, we visualize the representation of the feature distribution of our method and baseline on 10\% labeled data by using a scatter plot, as shown in Figure~\ref{fig:3}. We observe that the prediction results of our model can achieve better alignment with the ground truth compared to the baseline, and these results indicate that our method by introducing the reference network and confidence memory can provide more accurate pseudo-labels to supervise the student network.

In addition, we show the results of four test examples in Figure~\ref{fig:4}, and we show the predicted scores of the baseline, our proposed method, and ground truth. The baseline and our method's models are trained on 10\% of the labeled data.
As can be seen from the figure, the error between the predicted score of our method and the ground truth is smaller than the error between the baseline and the ground truth, which can prove the superiority of the assessment performance of our method.

\section{Conclusion}
In this paper, we propose a novel semi-supervised teacher-reference-student architecture for the AQA task, which can leverage both unlabeled and labeled data to improve the generalization of assessment. Specifically, 
the teacher network predicts pseudo-labels for unlabeled data thereby supervising the student network. The reference network provides additional supervision information for the student by referring to other actions.
Besides, the confidence memory is introduced to further enhance the accuracy of pseudo-labels generated by the teacher network and reference network.
Extensive experiments validate the effectiveness and superiority of our method over the other existing methods.

\section*{Acknowledgements}
This work is partly supported by the Funds for the Innovation Research Group Project of the NSFC under Grant 61921003, the NSFC Project under Grant 62202063, Beijing Natural Science Foundation (No.L243027)  and 111 Project under Grant B18008.

%
%
\bibliographystyle{splncs04}
\bibliography{main}

\end{document}